\documentclass{article}
\usepackage{paper_arxiv}
\pdfoutput=1
\usepackage[utf8]{inputenc}
\usepackage[T1]{fontenc}
\usepackage{hyperref}
\usepackage{url}
\usepackage{booktabs}
\usepackage{amsfonts}
\usepackage{nicefrac}
\usepackage{microtype}
\usepackage{lipsum}
\usepackage{graphicx}
\graphicspath{ {./images/} }

\usepackage{scalerel}
\usepackage{tikz}
\usetikzlibrary{svg.path}

\definecolor{orcidlogocol}{HTML}{A6CE39}
\tikzset{
  orcidlogo/.pic={
    \fill[orcidlogocol] svg{M256,128c0,70.7-57.3,128-128,128C57.3,256,0,198.7,0,128C0,57.3,57.3,0,128,0C198.7,0,256,57.3,256,128z};
    \fill[white] svg{M86.3,186.2H70.9V79.1h15.4v48.4V186.2z}
                 svg{M108.9,79.1h41.6c39.6,0,57,28.3,57,53.6c0,27.5-21.5,53.6-56.8,53.6h-41.8V79.1z M124.3,172.4h24.5c34.9,0,42.9-26.5,42.9-39.7c0-21.5-13.7-39.7-43.7-39.7h-23.7V172.4z}
                 svg{M88.7,56.8c0,5.5-4.5,10.1-10.1,10.1c-5.6,0-10.1-4.6-10.1-10.1c0-5.6,4.5-10.1,10.1-10.1C84.2,46.7,88.7,51.3,88.7,56.8z};
  }
}

\newcommand\orcidicon[1]{\href{https://orcid.org/#1}{\mbox{\scalerel*{
\begin{tikzpicture}[yscale=-1,transform shape]
\pic{orcidlogo};
\end{tikzpicture}
}{|}}}}

\usepackage{hyperref}

\title{OpenAI’s GPT4 as coding assistant}

\author{
 Lefteris Moussiades \orcidicon{0000-0003-0281-9231}\\
  Computer Science Department\\
  International Hellenic University\\
  Greece, Kavala PA 65404 \\
  \texttt{lmous@cs.ihu.gr} \\
   \And
 George Zografos \orcidicon{0009-0002-3417-049X}\\
  Computer Science Department\\
  International Hellenic University\\
  Greece, Kavala PA 65404\\
  \texttt{gezozra@cs.ihu.gr} \\
}

\usepackage{listings}
\usepackage{color}

\definecolor{dkgreen}{rgb}{0,0.6,0}
\definecolor{gray}{rgb}{0.5,0.5,0.5}
\definecolor{mauve}{rgb}{0.58,0,0.82}

\begin{document}
\maketitle
\begin{abstract}
Lately, Large Language Models have been widely used in code generation. GPT4 is considered the most potent Large Language Model from Openai. In this paper, we examine GPT3.5 and GPT4 as coding assistants. More specifically, we have constructed appropriate tests to check whether the two systems can a) answer typical questions that can arise during the code development, b) produce reliable code, and c) contribute to code debugging. The test results are impressive. The performance of GPT4 is outstanding and signals an increase in the productivity of programmers and the reorganization of software development procedures based on these new tools. 
\end{abstract}

\section{Introduction}
Among other features, Large Language Models (LLM) can generate code in various programming languages \cite{tamkin_understanding_2021}. Recently, many publications have recommended and evaluated LLMs specialized in code generation.

CodeBERT is a bimodal pre-trained model designed for programming and natural language tasks, like code search and documentation generation. It's developed using a Transformer-based architecture \cite{vaswani_attention_2023} and trained with a unique objective function to effectively use paired and unpaired data from programming and natural language sources \cite{feng_codebert:_2020}.

Codex is a GPT language model fine-tuned on public GitHub code, and a version of it powers GitHub Copilot. When evaluated on the HumanEval set, designed to gauge program synthesis from docstrings, Codex solves 28.8\% of the tasks, outperforming GPT-3 and GPT-J. The study also uncovers that multiple samplings from Codex enhance problem-solving success rates. Additionally, the paper discusses the challenges and broader implications of advanced code generation technologies \cite{chen_evaluating_2021}.

The capabilities of large language models in synthesizing Python programs from natural language prompts using two new benchmarks, MBPP and MathQA-Python, are explored by \cite{austin_program_2021}. The study reveals that as model size increases, synthesis performance also improves, with the largest models being able to correctly generate solutions to nearly 60\% of MBPP problems through few-shot learning. The models also benefit from human feedback, cutting error rates in half, but struggle to predict the outputs of the generated programs when provided with specific inputs.

Study \cite{lu_reacc:_2022} introduces a novel approach to code completion using an "external" context, emulating human behaviour of referencing related code snippets. The proposed framework combines retrieval techniques with traditional language models to better predict code, factoring in direct copying and semantically similar code references. When tested on Python and Java, this method achieves state-of-the-art performance on the CodeXGLUE benchmark.

Paper \cite{zan_cert:_2022} explores LLMs trained on unlabeled code corpora for code generation. It introduces CERT, a two-step method that creates a basic code outline and then fills in the details. The study also presents two new benchmarks, PandasEval and NumpyEval, for evaluating library-oriented code generation.

PanGu-Coder is a pre-trained language model built on the PanGu-Alpha architecture designed to generate code from natural language descriptions. The model is trained using a two-stage strategy, starting with raw programming data, followed by task-focused training using Causal and Masked Language Modelling objectives \cite{christopoulou_pangu-coder:_2022}

Li et al. introduced AlphaCode, a deep-learning model built with self-supervised learning and an encoder-decoder transformer, which approximates human-level performance in computer programming competitions on the Codeforces platform. Authors argue that this advancement could significantly boost programmers' productivity and reshape programming culture, where humans primarily define problems and machine learning handles code generation and execution \cite{li_competition-level_2022}.

CODEGEN is a family of large language models trained on natural language and programming data to advance program synthesis. The study also explores a multi-step approach to program synthesis, revealing improved performance when tasks are broken down into multiple prompts, and introduces an open benchmark, the Multi-Turn Programming Benchmark (MTPB), for this purpose \cite{nijkamp_codegen:_2022}.

Paper \cite{sandoval_lost_2022} investigates the impact of LLMs, like OpenAI Codex, on developers' code security. Through a user study involving 58 student programmers, the research examines the code's security when implementing a specific C-based task with the assistance of LLMs. The findings suggest that using LLMs does not substantially increase the risk of introducing critical security vulnerabilities in such coding tasks.

RepoCoder \cite{zhang_repocoder:_2023} is a framework designed for repository-level code completion that efficiently leverages information scattered across different files in a repository. RepoCoder uses a combination of a similarity-based retriever and a pre-trained code language model, along with an innovative iterative retrieval-generation approach, to improve code completion at various levels of granularity. RepoCoder has been tested on a new benchmark called RepoEval.

Paper \cite{zan_large_2022} thoroughly surveys 27 large language models geared explicitly towards the NL2Code task, which involves generating code from natural language descriptions. The study evaluates these models using the HumanEval benchmark and derives that success in this domain hinges on "Large Size, Premium Data, Expert Tuning". The authors also introduce a dedicated website to monitor ongoing advancements and discuss the gap between model performance and human capabilities in the NL2Code realm.

The BigCode community has unveiled StarCoder and StarCoderBase, advanced Large Language Models designed for code generation and infilling, with StarCoderBase trained on a vast dataset called The Stack and StarCoder being a fine-tuned version for Python \cite{li_starcoder:_2023}. 

WizardCoder is a model that empowers Code Large Language Models (Code LLMs) with complex instruction fine-tuning by adapting the Evol-Instruct method to the domain of code. It has been introduced in a paper \cite{luo_wizardcoder:_2023} and has demonstrated exceptional performance in code-related tasks.

Study \cite{xiao_supporting_2023} investigates the use of large language models (LLMs) to aid in deductive coding, a method in qualitative analysis where data is labelled based on predetermined codebooks. The approach reached satisfactory alignment with expert-labelled outcomes by integrating GPT-3 with expert-created codebooks for a specific task related to coding curiosity-driven questions. The paper highlights the potential and challenges of employing LLMs in qualitative data coding and broader applications.

One result of all this development is the addition of intelligent assistants to many well-known IDEs. For example, Visual Studio Code is supported by IntelliCode, PyCharm by Code With Me, Eclipse by Code Recommenders, NetBeans by Deep Learning, IntelliJ IDEA by Code With Me, and Xcode by SourceKit-LSP \cite{okeke_12_2022}.

In March 2023, Openai published the GPT-4 system card, which \cite{noauthor_gpt-4_nodate} analyzes the capabilities of GPT-4, including code generation. However, to date, we have not found any publication evaluating the coding capabilities of GPT-4.
This paper evaluates GPT-4 and GPT-3.5 as coding assistants.

\section{Methodology}
We consider three tasks for which a coding assistant should be helpful: Code development, Code Debugging, and answering questions related to code. Code development and Code debugging are self-explanatory concepts. The human programmer often has questions during code writing, such as details on the syntax of a command. For this reason, we check that GPT-3.5 and 4 can answer questions about the code satisfactorily.

There are many source code datasets, several mentioned in the introduction. However, these are geared to check LLMs' code production specifically. In addition, problems of a prototypical nature often arise in the production environment. Although we do not know exactly which data sets GPT-3.5 and 4 are trained on, it is reasonable to assume that they are trained on public data sets whose purpose is to evaluate LLMs' coding capabilities. For the reasons above, our tests do not rely on such data sets. Instead, we have carefully constructed 3 test suites: one for testing code generation capabilities, one for testing debugging capabilities, and one for answering questions. The tests were designed to limit the chances that GPT3.5 and 4 were trained on exactly those requested codes.
The tests were submitted through the web interface of GPT3.5 and 4. The prompt engineering of the tests follows the GPT best practices of  Openai \cite{noauthor_openai_nodate}. The results were evaluated based on an expert human reviewer or compared to another reliable source. As the tests are about checking different capabilities, more details about the test configuration and the evaluation of the results are given with the description of each test.
Java was used as the programming language.
All code and other answers generated by GPT3.5 and 4 is on GitHub [19]. 

\section{Answering questions}
In this task, we test the assistants to see if they can answer questions that often arise for developers when developing code. For this purpose, we constructed three questions of relative difficulty. We list the relevant prompts and then evaluate the assistants' answers.

\begin{itemize}
    \item \textit{Question 1 (Prompt)}:\\
    Does Java support passing a function as an argument to a function? What is the syntax?
\end{itemize}

\begin{itemize}
    \item \textit{Question 2 (Prompt)}:\\
    Consider the code

    System.out.print(s==s1+" "+s.equals(s1));

    I expected it to display two boolean values, but it displays only one. 
    Explain why?
\end{itemize}

\begin{itemize}
    \item \textit{Question 3 (Prompt)}:\\
    Non-abstract methods have an implementation. The same applies to the default methods.
    
    Non-abstract methods are inherited and can be overwritten. The same applies to default methods.
    
    What is the difference between default methods and non-abstract ones?
    Answer briefly.
\end{itemize}

\textbf{Response}

GPT3.5 and 4 responses were evaluated by a human expert and found to answer all three questions satisfactorily. Responses can be found on Github \cite{githubGitHubLmousGPT4-as-coding-assistant}.

\section{Code Development Assistance}
For code development, we constructed two tests. The first asks for developing a power function, and the second for implementing a tic-tac-toe application with predetermined classes.
\subsection{Power function (PF)}
In this task, we asked GPT3.5 and 4 to implement a function that calculates the power of a real number raised to an integer exponent. Although the task seems simple at first glance, it is demanding when high calculation precision is required. The difficulty arises from the approximate nature of real numbers. Due to the approximate nature of real numbers, the results of operations lack precision. When there are many intermediate operations, the deviations from each operation accumulate, and the final result may present a significant deviation. So, this is a complex implementation when precision is required in the calculations. Moreover, it is a feature, not a concern for application developers, as all languages provide a ready-made power function. Besides, after an exhaustive search on the web, we could not find a high-precision implementation.

\textbf{Evaluation}\\
The generated functions were compared with the Java Math.pow function. The Math.pow() function is implemented in Java as a native method, which means that it is implemented in the underlying platform’s native code. The implementation of Math.pow() varies depending on the platform and the underlying hardware architecture. The algorithm is optimized for speed and accuracy and is presumed to be relatively accurate.
The results were checked based on the following procedure.

Let GPT4.pow be the function produced by GPT4 and r(f,b,e) the result of the function f with base b and exponent e. For each b from 500 to 1000 with step 1 and each e from 0 to 9 with step 1, the values r(GPT4.pow,b,e) and r(Math.pow,b,e) are calculated. Assume that for each pair of these values, even one is non-infinite, and they differ from each other by more than 4.9E-324 (the smallest real value represented by Java double type). In that case, the absolute value of their difference is added to an appropriate adder. Then, the adder is divided by the number of terms in the sum and, thus, the average deviation of the GPT4.pow results from the Math.pow results are calculated. The same process is repeated to compare GPT3.5.pow to Math.pow.
The whole process is repeated for exponents from -1 to -9.\\

\textbf{{PF Prompt \#1}}

Develop a Java function that calculates the power of a real number raised to an integer exponent.

Specifications:

\begin{enumerate}
    \item Interface:  public static double pow(double b, int e)
    \item Don’t use Math.pow or BigDecimal.pow
    \item Achieve the maximum possible precision 
\end{enumerate}

\textbf{Response}
\newline Both systems responded by providing a satisfactory implementation based on the exponentiation by squaring algorithm. The algorithm has time complexity O(log n), where n is the exponent. The implementations are almost identical, with only two minor differences:

\begin{itemize}
    \item GPT4 checks if the exponent is odd by performing a bitwise and with 1 $((e \& 1)==1)$ while GPT3.5 performs an integer division remainder calculation $(e \% 2 == 1)$
\end{itemize}

\begin{itemize}
    \item GPT4 performs a right shift by 1 to divide the exponent by 2 $( e >>= 1)$, whereas GPT3.5 performs integer division $(e/=2)$ for the same purpose.
\end{itemize}

The algorithms presented the same average deviation with respect to Math.pow, which was 2.356527240763158E10 for positive exponents and 1.7112490986192953E-22 for negative exponents.\\

\textbf{PF Prompt \#2}

Can you improve the precision of your function? I checked it against Math.pow and found significant discrepancies.

Examples:

base = 502, exponent= 9, GPT.pow = 2.0245730632526733E24, Math.pow = 2.024573063252673E24, diference = 2.68435456E8
\\base = 504, exponent = 9, GPT.pow = 2.098335016107156E24, Math.pow = 2.0983350161071556E24, diference = 2.68435456E8\\

\textbf{Response}

GPT3.5 responded with a function that implements the Taylor series expansion \cite{wolframTaylorSeries} algorithm, which increases time complexity to O(e2). GPT4 again used exponentiation by squaring but used the BigDecimal class \cite{noauthor_bigdecimal_nodate}, recommended for cases requiring precision in calculations.

The mean deviation of GPT3.5 worsened to 2.2292150579952536E25 for positive exponents and 1.0012331308931004 for negative ones.

The mean deviation of GPT4 improved to 2.3037066373333335E9 for positive exponents and 2.1726446876877912E-2 for negative ones.  

\subsection{Tic-Tac-Toe application (TTT)}

In this task, we asked GPT to develop a tic-tac-toe application following especial specifications. We set certain specifications to minimize the chance that a tic-tac-toe app would be found ready-made and delivered intact.\\

\textbf{TTT Prompt \#1}

Develop a command-line tic-tac-toe application consisting of the following classes:

Player, Board, LivePlayer, RBPlayer, and Game.

\begin{itemize}
    \item \textit{Player}: Is an Abstract class containing , final char id, abstract method Board move(Board board)
\end{itemize}

\begin{itemize}
    \item \textit{Class Board}: Represents the game board. It contains the following public function members: 
    \\void displayBoard(): It displays the game board on its current status 
    \\char win(): It returns the winner’s id. If there is no winner, it returns a white character.
\end{itemize}

\begin{itemize}
    \item \textit{Class LivePlayer}: Represents a human player. It is a concrete class implementation inherited from Player.
\end{itemize}

\begin{itemize}
    \item \textit{Class RBPlayer}: Represents an artificial Rule-based Player. It is based on the following rules:  
    \\A. If there is a movement to win, select it.  
    \\B. If the opponent has a movement to win, select it to block the opponent from winning.
\end{itemize}

\begin{itemize}
    \item \textit{Game}: Uses the above-described classes to implement a tic-tac-toe game.
\end{itemize}

\textbf{Response}

GPT4 respond with a fully functional application that meets all our requirements. The code quality is good, including a warning that the used Board object could have been declared final.

GPT3.5 responded with code that contained compile time errors. We performed the following communication to investigate its ability to produce correct code.\\

\textbf{TTT Prompt \#1.1}

Your code compiles with errors. 
Examples:
\begin{itemize}
    \item error: cells has private access in Board
    \\board.cells[i][j] = id;
\end{itemize}
\begin{itemize}
    \item error: cannot assign a value to final variable board
    \\board = currentPlayer.move(board);
\end{itemize}
Rewrite code to avoid compile-time errors.

GPT3.5 replied with code containing logical errors. We prompt it as follows:\\

\textbf{TTT Prompt \#1.2}

Your code has logical errors.
Here is the output of your code after two movements of each player

Player X, enter your move (row [0-2] and column [0-2]): \textbf{1 1}
\begin{center}
    -------------
    \\|\hspace{10pt}|\hspace{10pt}|\hspace{10pt}| 
    \\-------------
    \\|\hspace{10pt}|\hspace{2pt}X\hspace{2pt}|\hspace{10pt}| 
    \\-------------
    \\|\hspace{10pt}|\hspace{10pt}|\hspace{10pt}| 
    \\-------------
    
    -------------
    \\|\hspace{2pt}O\hspace{2pt}|\hspace{10pt}|\hspace{10pt}| 
    \\-------------
    \\|\hspace{10pt}|\hspace{10pt}|\hspace{10pt}| 
    \\-------------
    \\|\hspace{10pt}|\hspace{10pt}|\hspace{10pt}| 
    \\-------------
\end{center}

After the second fix, in the third version of the application, GPT3.5 responded with functional code.

GPT4 respond with a fully functional application that meets all our requirements. The code quality is good, including a warning that the used Board object could have been declared final.

Next, we requested a new class representing an artificial player based on the minimax \cite{75b5cec4-4f40-3f36-9492-860b8376add8} algorithm. The minimax implements a perfect player, i.e., a player who never loses. Therefore, the worst possible outcome minimax may give is a draw.\\

\textbf{TTT Prompt \#2}

Can you add the class MinimaxPlayer representing an artificial player based on the well-known minimax algorithm?

\textbf{Response}

GPT4 responded with a fully functional minimax player. GPT3.5 replayed with an erroneous version of a minimax player. A communication ensued in which we attempted to inform GPT3.5 of its errors, but it failed to present a satisfactory solution. Finally, we prompt GPT3.5 as follows:\\

\textbf{TTT Prompt \#2.1}

No improvement. It's still straightforward for anyone to win your MinimaxPlayer. I'm giving you the game board if it can help you. Please don't give me the same wrong algorithm again. If you can't do better, just let me know.

Player X, enter your move (row [0-2] and column [0-2]): \textbf{2 0}

\begin{center}
    -------------
    \\|\hspace{2pt}O\hspace{2pt}|\hspace{2pt}O\hspace{2pt}|\hspace{2pt}X\hspace{2pt}| 
    \\-------------
    \\|\hspace{2pt}O\hspace{2pt}|\hspace{2pt}X\hspace{2pt}|\hspace{10pt}| 
    \\-------------
    \\|\hspace{2pt}X\hspace{2pt}|\hspace{10pt}|\hspace{10pt}| 
    \\-------------
    
\end{center}

Player X wins!

Here, GPT3.5 explained the difficulties of implementing the algorithm and suggested that we study the matter more or look for a ready-made solution on GitHub.

\section{Debugging Assistance (DA)}

To test the debugging capabilities, we designed two tests. One includes code that throws an exception, and the other includes code containing a logic error. 

\subsection{Exception (E)}

In this task, we provided a code that crashes with IndexOutOfBoundsException and asked GPT3.5 and 4 to explain the problem and fix the code.\\

\textbf{DA-E Prompt \#1}

The Code below fails with IndexOutOfBoundsException.

\begin{lstlisting}
import java.util.ArrayList;
import java.util.List;

public class Debug2 {
    static ArrayList<String> l=new ArrayList<>();
    
    static void load() {
        l.add("Green");
        l.add("Black");
        l.add("Blue");
        l.add("White");
        l.add("Pink");
        l.add("Black");
    }
    
    static void delAll(List<String> l, String target) {
        int size=l.size();
        for (int i=0; i<size; i++)
            if (target.equals(l.get(i))) {
                l.remove(i);
            }
    }
    
    public static void main(String[] args) {
        load();
        delAll(l,"Black");
    }
}

\end{lstlisting}

Explain the error and correct the code.\\

\textbf{Explanation of the error}

First, the exception is raised in the delAll function, which is responsible for deleting all the target elements from the list l. The function stores the list size in the local variable size and then, in the iterative process, tries to delete every element equal to the target. However, after deleting the first element, the list size is reduced by 1. However, delAll tries to access the list for its original size, which leads to the exception.

\textbf{Responce}

Both assistants solved the problem successfully. While GPT3.5 proposed a solution based on an Iterator, GPT4 proposed two alternatives. In the first solution, the for control expression replaces the size variable with the function that returns the list size (l.size()); inside the for, decrements i by one each time it deletes an element. The second solution traverses the list from the end (l.size()-1) to the beginning, thus ensuring no IndexOutOfBoundsException issue. 

\subsection{Logical Error (LE)}

\textbf{DA-LE Prompt \#1}

The code below contains logical errors.

Expected Output: [1, 2, 3, 4, 0, 5, 6]
\\Actual Output: [1, 2, 3, 4, 5, 6, 0, 0, 0, 0]

Explain the errors and correct the code.

\begin{lstlisting}
// Code containing logical error

import java.util.Arrays;

public class Debugging {

    static int[] resize(int[] input, int newSize) {
        return Arrays.copyOf(input, newSize < input.length ? newSize : input.length);
    }

    static int add(int[] array, int data, int index) {
        for (int i = 0; i <= index; i++) {
            if (array[i] == data) {
                return index;
            }
        }
        array[index++] = data;
        return index;
    }

    static int[] generateSet(int... array) {
        int[] set = new int[array.length];
        int idx = 0;
        for (int element : array) {
            idx = add(set, element, idx);
        }
        resize(set, idx);
        return set;
    }

    static int[] concat(int[] array1, int[] array2) {
        int[] rslt = new int[array1.length + array2.length];
        System.arraycopy(array1, 0, rslt, 0, array1.length);
        System.arraycopy(array2, 0, rslt, array1.length, array2.length);
        return generateSet(rslt);
    }

    public static void main(String[] args) {
        int[] set1 = generateSet(1, 2, 3, 4, 0),
                set2 = generateSet(0, 3, 4, 5, 6);
        int[] union = concat(set1, set2);
        System.out.println(Arrays.toString(union));
    }
}

\end{lstlisting}

\textbf{Explanation of the error }

There are two bugs in the code. The first one is found in generateSet, which calls the function resize but does not assign the array returned by resize to the set variable. Thus, the set retains its original size and data. So the fix needed here is return resize(set, idx); instead of resize(set,idx); return set;
The second error is within the add function, which iterates while i<=index, whereas the correct condition is i<index.

\textbf{Response}

First, GPT3.5 and GPT4 correctly explained the problems in the add and generateSet functions. In addition, they identified a resize problem when there is none. More specifically, GPT3.5 commented:

\begin{enumerate}
    \item The resize method is not updating the size of the array correctly. It creates a new array of the specified size but doesn't copy the elements from the original array.
    \item Use Arrays.copyOf to create a new array of the desired size and copy the elements from the original array to the new one.
\end{enumerate}

And GPT4 commented:

\begin{enumerate}
    \item Resize method: In the current implementation, if newSize is larger than input.length, it would return an array of the same size as input. This does not match the intended behavior of resizing the array to newSize.
\end{enumerate}

These comments are wrong.

However, the generated codes are functional as they correctly fix both add and generateSet, while the change they make to resize does not affect the specific code. More specifically, both systems converted resize so that it does not support reducing the size of the input table. Indeed, size reduction is not needed in this code. Of course, a resize that helps reduce a table's length (with possible data loss) might be helpful elsewhere.

\section{Conclusions}
In this work, we examined the potential of GPT3.5 and 4 as coding assistants for three distinct tasks: Answering questions and providing Development and Debugging assistance. In answering questions, both LLMs proved to be efficient. In Development assistance, GPT4 proved superior to GPT3.5. Both in creating the pow function, it achieved a significant improvement in accuracy, and in the requirements for the tic-tac-toe application, it immediately responded with complete success. Moreover, it added a player based on the Minimax algorithm with ease. This is a requirement, according to our estimation, that is far from easy to implement. GPT3.5 failed to meet this requirement. In testing the debugging capabilities, GPT3.5 and 4 responded promptly and successfully to exception and logical error investigations. 
These conclude that GPT4 can provide substantial and reliable help as a coding assistant for all three properties tested. As expected, GPT3.5 appeared inferior to GPT4, but its capabilities are still impressive.
Recently, a heated debate has been about whether artificial intelligence will replace human programmers. We believe the answer to this question is impossible, as no one can predict the future. However, currently, GPT4 can provide meaningful and reliable assistance to coding and dramatically improve the productivity of human developers. Such a thing is sure to reorganize the software production processes and possibly will not leave the job market of programmers unaffected.
Whether its effect will increase the amount of software produced or unemployment in the developer industry remains to be seen. 

\bibliographystyle{ieeetr}
\bibliography{paper_arxiv}
\end{document}